\documentclass{article}

\PassOptionsToPackage{numbers, compress}{natbib}

\usepackage[preprint]{nips_2018} %

\usepackage[utf8]{inputenc} %
\usepackage[T1]{fontenc}    %
\usepackage{hyperref}       %
\usepackage{url}            %
\usepackage{booktabs}       %
\usepackage{amsfonts}       %
\usepackage{nicefrac}       %
\usepackage{microtype}      %

\usepackage{caption}

\usepackage{amsmath}
\DeclareMathOperator*{\argmax}{arg\,max}

\usepackage{graphicx}
\graphicspath{{figures/}}

\usepackage{caption}
\usepackage{subcaption}

\title{Training an Interactive Helper}

\author{
  Mark Woodward, Chelsea Finn, Karol Hausman\\
  Google Brain\\
  \texttt{markwoodward,chelseaf,karolhausman@google.com}\\
}

\begin{document}
\suppressfloats

\maketitle

\begin{abstract}

Developing agents that can quickly adapt their behavior to new tasks remains a challenge.
Meta-learning has been applied to this problem, but previous methods require either specifying a reward function which can be tedious or providing demonstrations which can be inefficient. %
In this paper, we investigate if, and how, a "helper" agent can be trained to interactively adapt their behavior to maximize the reward of another agent, whom we call the "prime" agent, without observing their reward or receiving explicit demonstrations.
To this end, we propose to meta-learn a helper agent along with a prime agent, who, during training, observes the reward function and serves as a surrogate for a human prime.
We introduce a distribution of multi-agent cooperative foraging tasks, in which only the prime agent knows the objects that should be collected.
We demonstrate that, from the emerged physical communication, the trained helper rapidly infers and collects the correct objects.
\end{abstract}

\section{Introduction}

Training reinforcement learning (RL) agents often requires millions of steps on the target task~\cite{duan2016bdr,kalashnikov2018qsd}.
While work is progressing on more sample-efficient algorithms~\cite{gu20162016cdq,chebotar2017cmm}, more progress is needed before agents can learn efficiently.

If we restrict the target tasks to be \emph{near} a distribution of training tasks, then we can apply meta-learning to train an agent to rapidly adapt to a task~\cite{schmidhuber1987eps,bengio1991lsl}. 
As an example, for a straightening robot, the target task might consist of the desired destinations of toys, dishes, and clothing for a specific home, and the training distribution would consists of a set of such tasks.

In this paper, we make use of meta-learning and focus on task distributions where the reward function varies across tasks, or equivalently where the desired behavior varies across tasks and is optimal under a corresponding reward function; e.g. tasks for the straightening robot. 
This is distinct from task distributions in which the reward function is constant across tasks but the environment varies, such as maze solving~\cite{jaderberg2017rlu}.

One class of meta-learning techniques for adapting to reward-varying tasks adapts the policy to rewards received during the task~\cite{finn2017mam,duan2016rl2,levine2018rlc}.
Unfortunately, these techniques require a reward function for the target task and specifying a reward function is often tedious, particularly as the task complexity grows.

Another class of meta-learning methods for adapting to reward-varying tasks adapts the policy to mimic expert demonstrations of the target task~\cite{yu2018osi,james2018tec}.
Unfortunately, providing demonstrations can be inefficient. %
For example, without interaction it is not always obvious what the agent is unclear about and what type of a demonstration, how complete of a demonstration, or how many demonstrations to give.

In this paper, we investigate if it is possible to train an agent to interactively maximize another agent's reward, without access to the reward function and without explicitly receiving demonstrations. We call the agent whose reward function is being maximized the "prime" agent, suggesting that it is the primary agent, and we call the other agent the "helper" agent.

In our proposed method, a helper agent is simultaneously meta-learned with a prime agent, who serves as a surrogate for a human prime.
By making the prime agent aware of their joint task-specific reward function, communication emerges allowing the helper to rapidly infer the reward function and adapt to the task.

Our work is most similar to that of~\citet{mordatch2018egc}.
While they address a different task distribution, the main difference is the asymmetry of knowledge and roles in our prime-helper tasks.

Our primary contribution is a method for training a helper agent to, through interaction, rapidly adapt its policy to maximize another agent's reward. We evaluate our method on simulated foraging tasks. Our preliminary results demonstrate that 1) the helper infers the task from a handful of prime agent actions and in some cases zero prime actions (which is itself a communication), 2) the prime agent receives more reward with the helper agent than without, and 3) the emerged communication and the prime agent's delegation to the helper agent is explicit and intentional.

\section{Methodology} \label{sec:methodology}
\begin{figure}
	\begin{center}
		\includegraphics[height=2in]{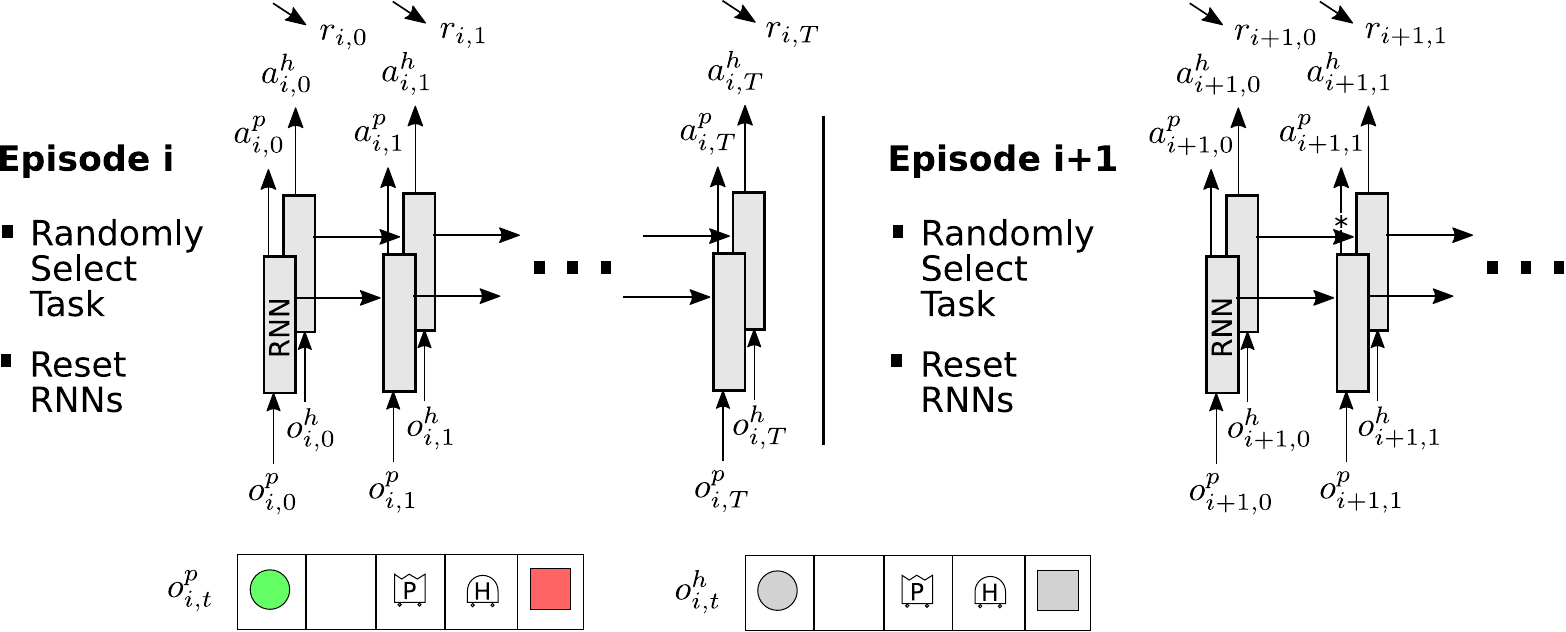}
	\end{center}
	\caption{
Training episode $i$ begins by randomly selecting a task and resetting the recurrent state for the prime ($p$) and helper ($h$) agents. On every step, $t$, of the episode, the agents receive their respective observations, $o^{p}_{i,t}$ and $o^h_{i,t}$, and select their respective actions, $a^{p}_{i,t}$ and $o^h_{i,t}$, and a joint reward is stored, $r_{i,t}$. The prime agent's observation, $o^{p}_{i,t}$, informs it of the the task for the episode, depicted with color in the gridworld example. At the end of the episode, the stored rewards are used to improve the prime and helper policies.
	}
	\label{fig:method}
\end{figure}

Similar to recent work in meta-learning, we train agents whose policies are functions of a recurrent neural network~\cite{duan2016rl2}.\footnote{This technique is commonly described as making use of a recurrent network but the main characteristic is that the policy is modified because of experience and not a change in network weights. For example attention over observations could also be used~\cite{sukhbaatar2015eem,vaswani2017aay}.}
Figure~\ref{fig:method} illustrates our proposed training method.
Each episode consists of the helper agent and the prime agent acting and receiving agent specific observations in a task drawn from a task distribution. 
During an episode, the policy weights are fixed and actions are a function of the hidden states.
At the end of each training episode, the policy weights of both agents are moved in a direction that would have increased the prime's reward for that episode's task. 
Key to our method is that, during the episode, only the prime has knowledge of the task's reward function.
Since the prime is the only agent with knowledge of the reward function yet the helper is available to help, communication emerges, and the helper is trained to infer the task rewards from the prime and adapt its policy to contribute to the prime's reward.

We simultaneously train the prime and the helper specifically to allow communication to emerge. By emerging the communication, we allow the agents to communicate in ways that a programmer may not have thought of. This is particularly true when optimal performance requires the prime to maintain beliefs about the helper. 
As an example, in our experiments, we noticed the prime occasionally communicating after sub-optimal helper behavior indicating that the prime was monitoring the helper's performance.

\section{Experiment: 1-D Foraging} \label{sec:experiment}

\subsection{Task} \label{sec:task}

\begin{figure}
	\begin{center}
		\includegraphics[width=5.5in]{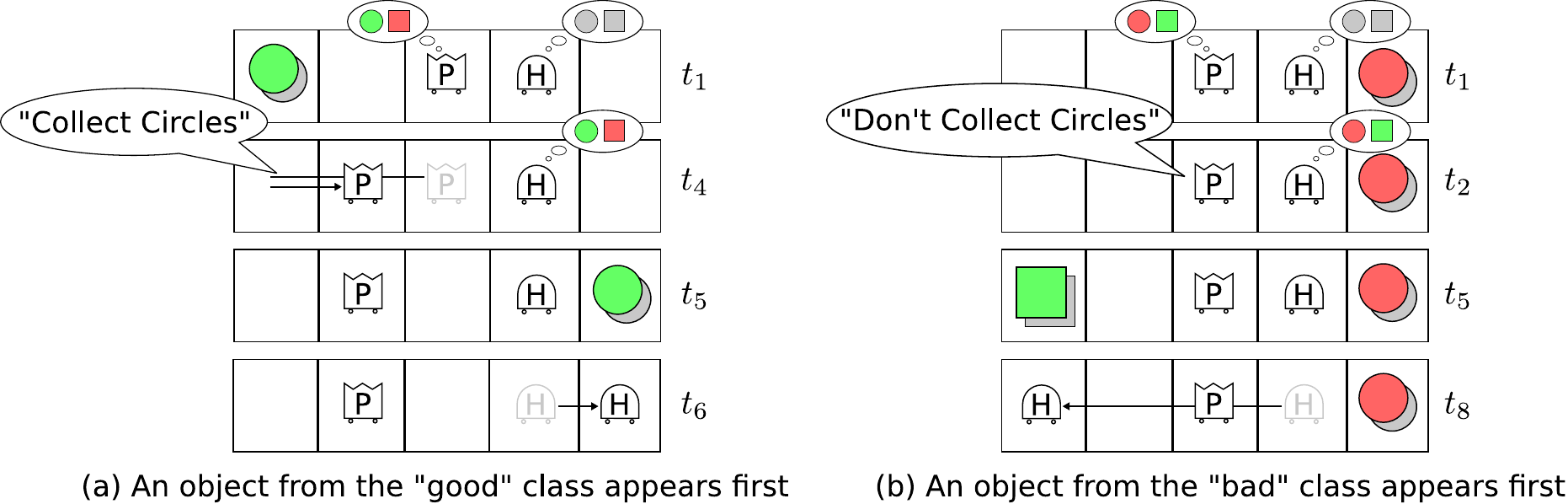}
	\end{center}
	\caption{
Two agents collect "good" objects and avoid "bad" objects in a 5 cell gridworld. The prime agent observes which objects are good and which are bad, depicted in color above the gray objects, and must communicate this to the helper agent. Learned policy: (a) if the first object to appear in the episode is a good object, then the prime agent collects that object, (b) if the first object to appear in the episode is a bad object, then the prime agent does not move. In both cases the helper infers the task and collects all future good objects.
	}
	\label{fig:communication}
\end{figure}

Figure~\ref{fig:communication} depicts two tasks drawn from our distribution of "foraging" tasks. Each task consists of the helper and prime located on a one dimensional gridworld, consisting of five cells. 
There are three actions for each agents: move left, move right, and don't move. 
Objects randomly drawn from two object classes appear alternately in the right-most and left-most cells, persist for 9 time steps, and then disappear. 
An observation, $o_t$, consists of a set of binary strings, one for each cell. 
A cell's binary string is of length 5 for the helper and 6 for the prime, with one bit for the presence of the prime, one bit for the presence of the helper, one bit for the presence of an object, two bits for the class of the object, and, for the prime, one bit indicating whether it is good to pick up objects of this class.
If an action, including no action, moves an agent to a cell that contained an object then the object disappears and a reward is stored for that timestep; $+1$ is stored for collecting a good object and $-1$ for collecting a bad object. 
Finally, a reward of $-0.1$ is stored for all timesteps in which the prime moves. 
This last penalty is added to encourage minimal prime movement, so that the ability to communicate and delegate to the helper is clear in the results.
Each episode runs for $100$ time steps, with exactly $20$ objects appearing in an episode.

As a baseline, we also train a prime agent acting alone without a helper agent, all other aspects of the model and training are identical.

\subsection{Model} \label{sec:model}

We use identical, but independent, deep recurrent Q-networks as the policy for each agent~\cite{mnih2013pad,hausknecht2015drq}. A single 200 unit LSTM is used as the recurrent network~\cite{hochreiter1997lst}. For each agent, on each timestep, their observation, $o_t$, described in section~\ref{sec:task}, is flattened and fed to the LSTM, and a linear layer is applied to the LSTM output, $h_t$, resulting in a 3 dimensional vector, $q_t$, which estimates the expected future reward for each action.
\begin{equation}
h_t = LSTM(o_t,h_{t-1})
\end{equation}
and
\begin{equation}
q_t = Wh_t+b,
\end{equation}
where $LSTM(o_t,h_{t-1})$ represents the LSTM update equations~\cite{hochreiter1997lst}. Either $\argmax(q_t)$ is chosen for $a_t$, or, during training, a random action is chosen with probability $0.05$ ($\epsilon$-greedy exploration).

At the end of each episode a gradient step is taken for each agent, using Adam with the default parameters, on the Q-Learning objective function\cite{kingma2015ams}:
\begin{equation}
  \mathcal{L}(\Theta) := \sum_t[q_t\cdot{\vec{a}}_t - (r_t + \gamma \max{}q_{t+1})]^2,
\end{equation}
where $\Theta$ are the parameters of the LSTM and linear layer, one set for each agent, and $\vec{a}_t$ is $a_t$ in one-hot form. We set $\gamma=0.95$ and we do not use a target network. We use a batch size of 100 episodes for each gradient update, and we perform 10,000 gradient updates.

\subsection{Results} \label{sec:results}

\begin{figure}
  \begin{minipage}[t]{2.64in}
    \centering
	\includegraphics[width=1\textwidth]{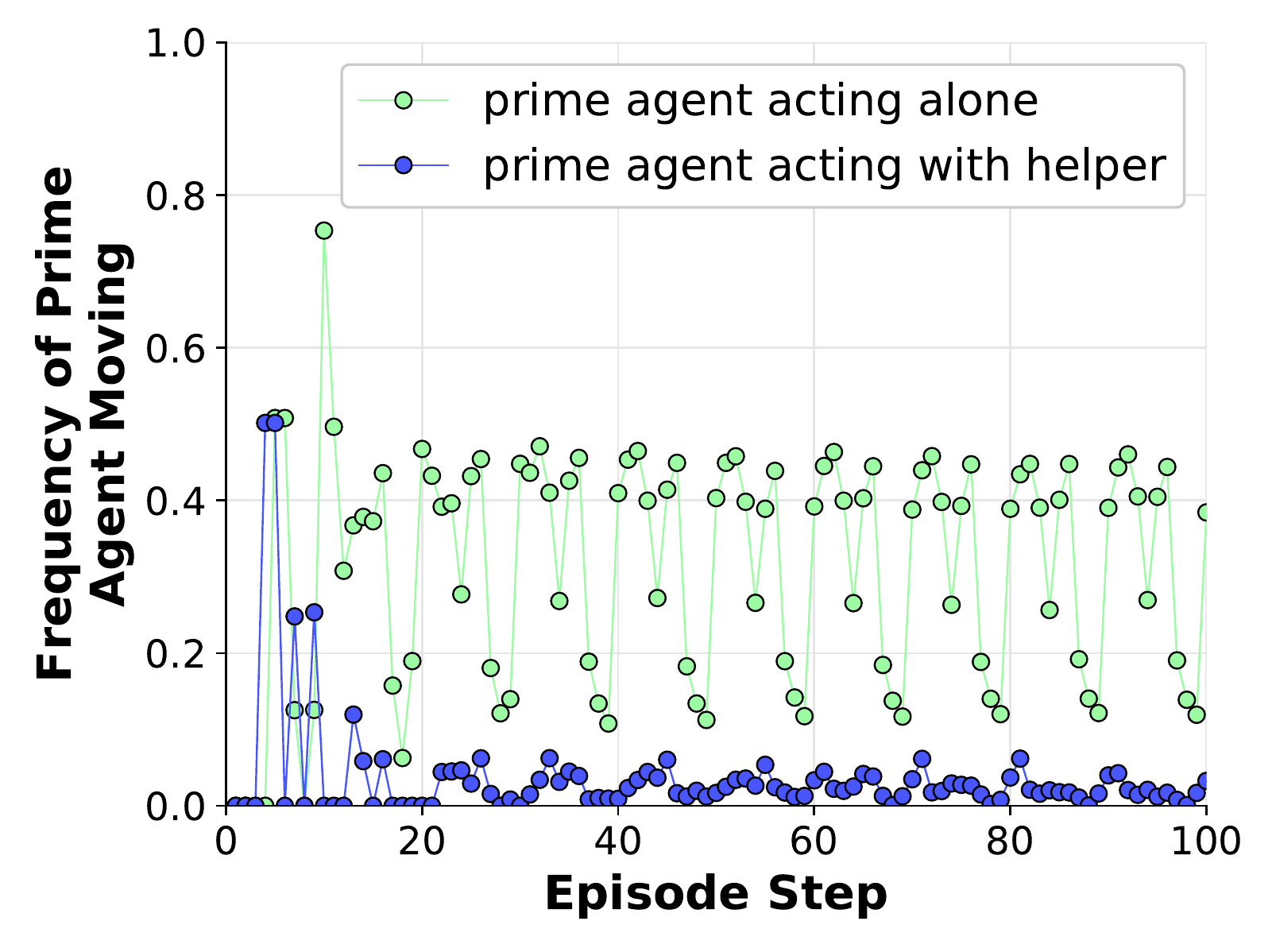}
  	\caption{
The prime agent moves less and mostly at the start of an episode when trained with a helper. The aliasing is due to the regular appearance of objects in an episode.
	}
	\label{fig:episode_actions}
  \end{minipage}
  \hfill
  \begin{minipage}[t]{2.64in}
    \centering
	\includegraphics[width=1\textwidth]{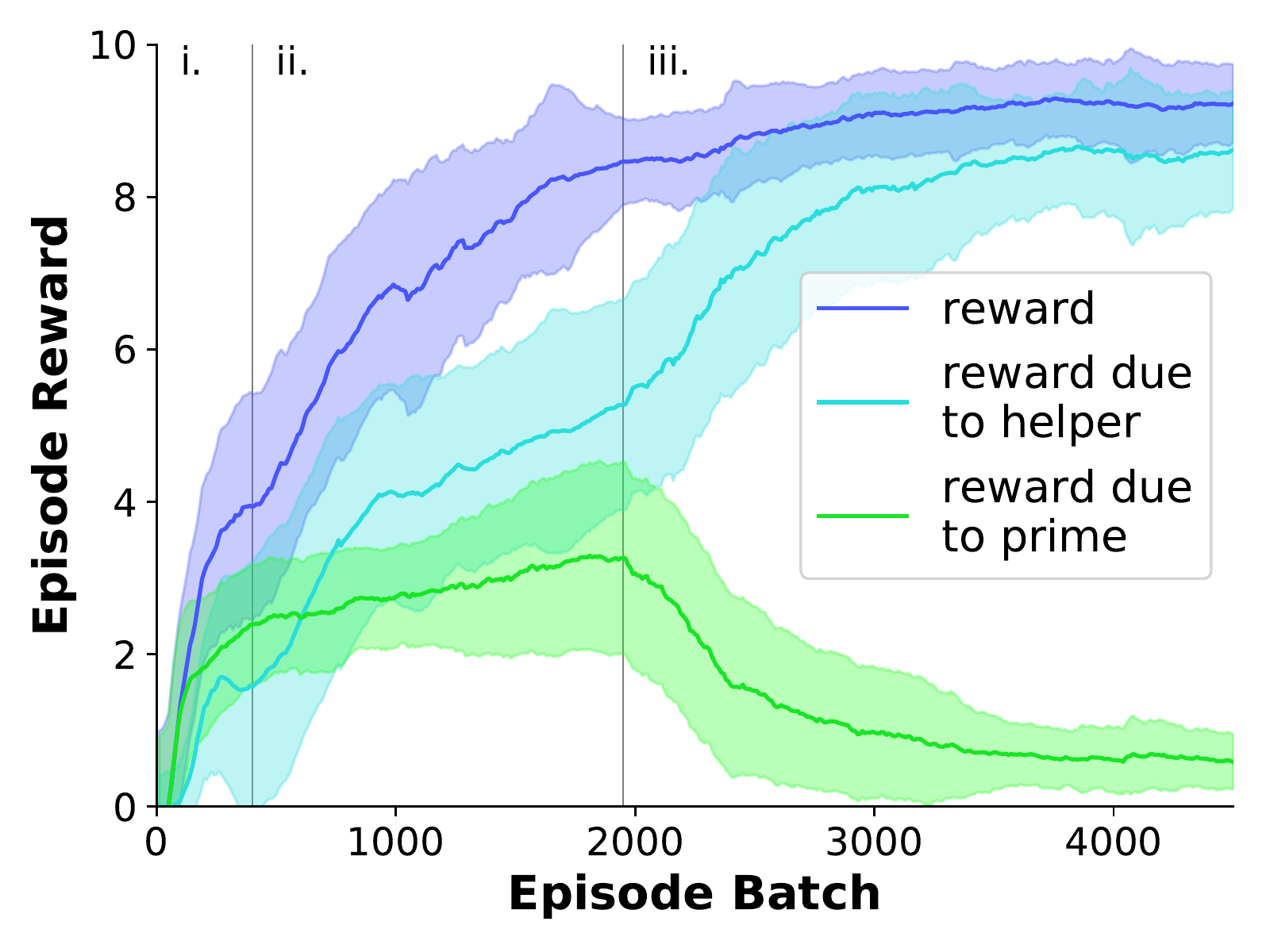}
  	\caption{
The episode reward due to the prime agent's actions rises, peaks, and then drops during training as the agents learn to communicate and delegate tasks to the helper.
	}
	\label{fig:episode_rewards}
  \end{minipage}
\end{figure}

The baseline prime, which acts alone, takes an average of $29.48$ actions per episode and receives an average reward of $5.99$ per episode out of a maximum expected reward of $10$, compared with $3.98$ actions per episode and an average reward of $9.29$ when assisted by a helper\footnote{20 objects appear per episode, with 10 being good objects on average, resulting in $+10$ reward if all good objects are collected and the prime does not move}.
Thus, the helper is helping the prime.
See figure~\ref{fig:episode_actions} for a plot of the prime's movement during an episode vs. the baseline prime.

Figure~\ref{fig:communication} shows the learned policies, which we constructed by observing trained agents.
If the first object to appear is a "good" object, then the prime collects the object, otherwise it does not move.
In either case, the helper collects all future good objects, even inferring from the prime's lack of motion on an initial bad object that the other object class is good.

Figure~\ref{fig:episode_actions} shows that, when assisted by the helper, the prime acts early in the episode and is largely dormant thereafter.
This, coupled with the observed policies of figure~\ref{fig:communication} and the high reward due to the helper's actions of figure~\ref{fig:episode_rewards}, demonstrates that the helper quickly infers and adapts to the task.

The emerged communication and policies demonstrate that the prime agent intentionally delegates to the helper. The delegation can be seen by the prime's lack of movement late in episodes, figure~\ref{fig:episode_actions}, and the prime's low contribution to the total reward, figure~\ref{fig:episode_rewards}. Further, even when a good object is next to the prime, it will wait for the helper to cross the grid rather than collecting it itself.

Lastly, we note that the agents move through three phases during training, as seen in figure~\ref{fig:episode_rewards}: i) first the prime learns to collect good objects, ii) then the prime and the helper learn to communicate and jointly collect good objects, iii) finally the agents learn to have the helper collect the good objects.

\section{Conclusion and Future Work} \label{sec:conclusion}

We presented a method for training a helper agent to maximize the reward of another agent, without requiring a reward function or explicit demonstrations for the target task. 
The results demonstrated that the helper agent adapts quickly and provides a net benefit to the prime agent.
We are currently working to achieve the optimal joint policy of the prime agent taking a single step when the first object to appear is a good object, instead of collecting the object.
We are also working on more complex tasks, including a 3D environment with first person pixel observations.

\bibliography{main}

\end{document}